\documentclass{article}
\usepackage{style/ijcai21}

\usepackage{times}
\usepackage{soul}
\usepackage{url}
\usepackage[hidelinks]{hyperref}
\usepackage[utf8]{inputenc}
\usepackage[small]{caption}
\usepackage{graphicx}
\usepackage{amsmath}
\usepackage{amsthm}
\usepackage{booktabs}
\usepackage{algorithm}
\usepackage{algorithmic}
\usepackage{bbm}
\usepackage{bbold}
\usepackage{multirow}
\usepackage{amsfonts}
\usepackage{subcaption}
\usepackage{bm}
\usepackage{amssymb}
\usepackage{pifont}
\usepackage{graphicx}
\usepackage{tabularx}

\usepackage{color}

\usepackage{latexsym}

\begin{document}


\title{Masked Contrastive Learning for Anomaly Detection}

\author{
    Hyunsoo Cho \qquad Jinseok Seol \qquad Sang-goo Lee \\
    Department of Computer Science and Engineering \\
    Seoul National University \\
    {\tt\small \{johyunsoo, jamie, sglee\}@europa.snu.ac.kr}
}

\maketitle

\begin{abstract}

    Detecting anomalies is one fundamental aspect of a safety-critical software system, however, it remains a long-standing problem.
    Numerous branches of works have been proposed to alleviate the complication and have demonstrated their efficiencies.
    In particular, self-supervised learning based methods are spurring interest due to their capability of learning diverse representations without additional labels.
    Among self-supervised learning tactics, \textit{contrastive learning} is one specific framework validating their superiority in various fields, including anomaly detection.
    However, the primary objective of contrastive learning is to learn \textit{task-agnostic} features without any labels, which is not entirely suited to discern anomalies.
    In this paper, we propose a \textit{task-specific} variant of contrastive learning named \textit{masked contrastive learning}, which is more befitted for anomaly detection.
    Moreover, we propose a new inference method dubbed \textit{self-ensemble inference} that further boosts performance by leveraging the ability learned through auxiliary self-supervision tasks.
    By combining our models, we can outperform previous state-of-the-art methods by a significant margin on various benchmark datasets.

\end{abstract}


\section{Introduction}

\begin{figure}[t!]
    \begin{subfigure}{.49\linewidth}
        \centering
        \frame{\includegraphics[width=.95\linewidth,height=.9\linewidth]{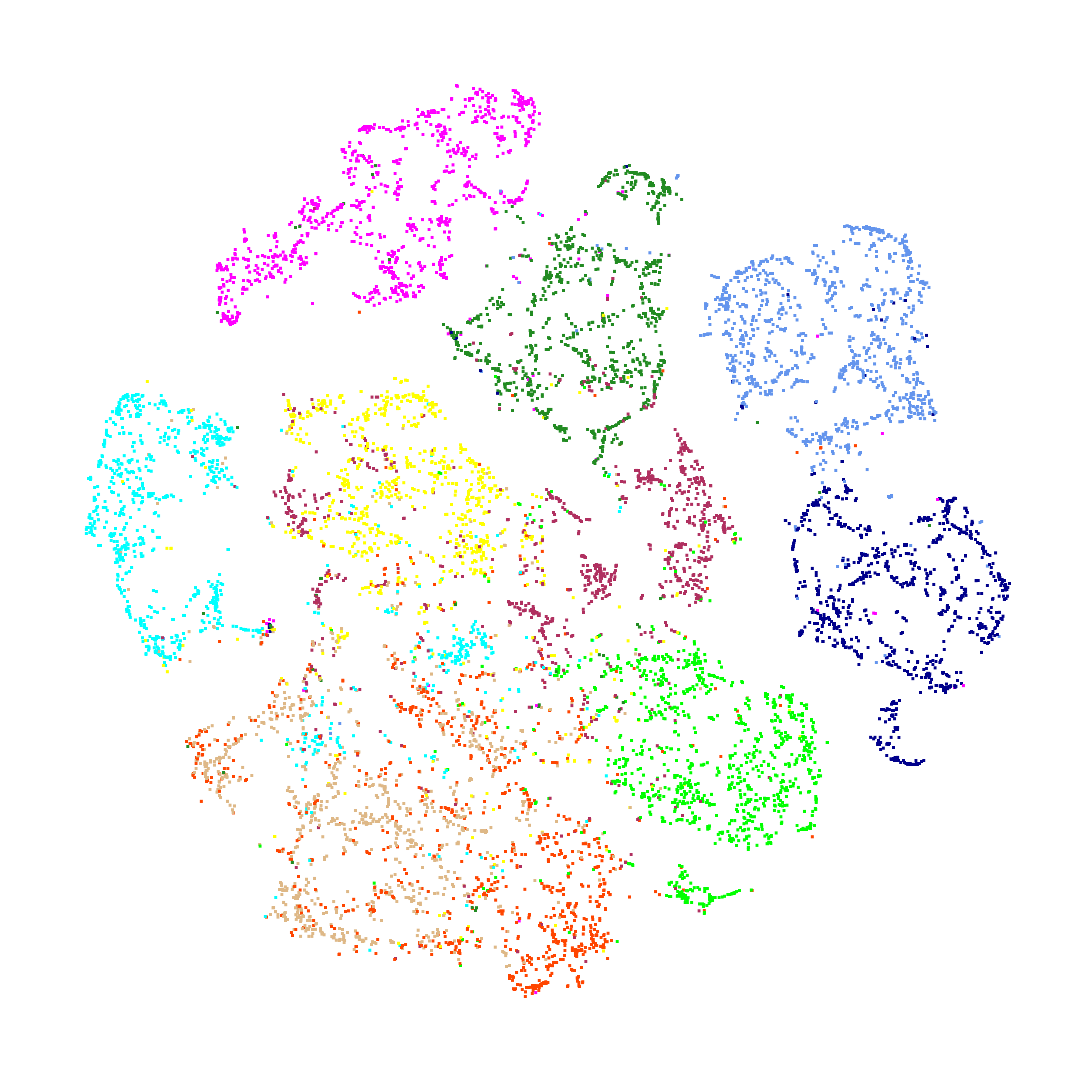}}
        \caption{SimCLR}
        \label{fig:tsne-simclr}
    \end{subfigure}
    \begin{subfigure}{.49\linewidth}
        \centering
        \frame{\includegraphics[width=.95\linewidth,height=.9\linewidth]{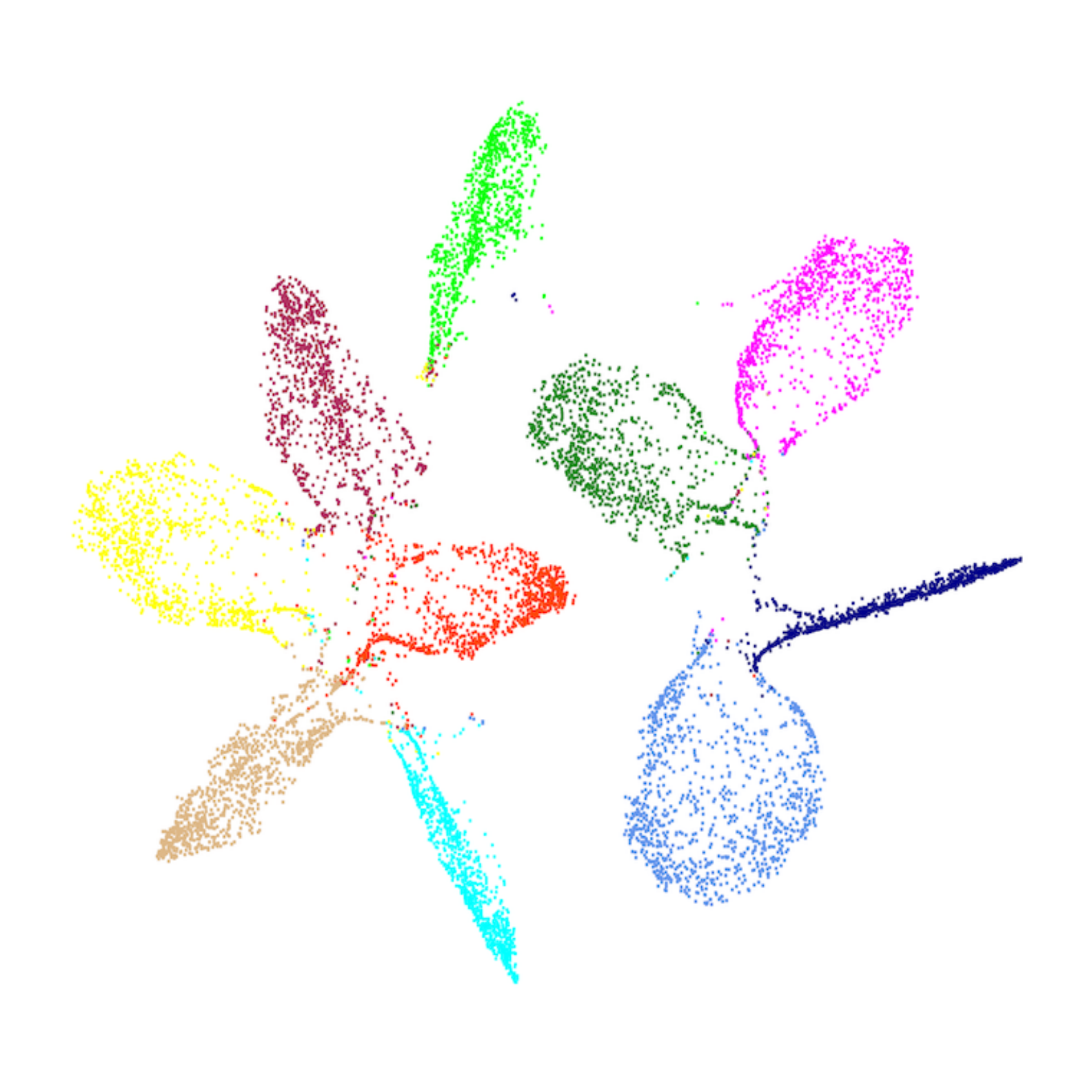}}
        \caption{SupCLR}
        \label{fig:tsne-supclr}
    \end{subfigure}
    \par \medskip
    \begin{subfigure}{.49\linewidth}
        \centering
        \frame{\includegraphics[width=.95\linewidth,height=.9\linewidth]{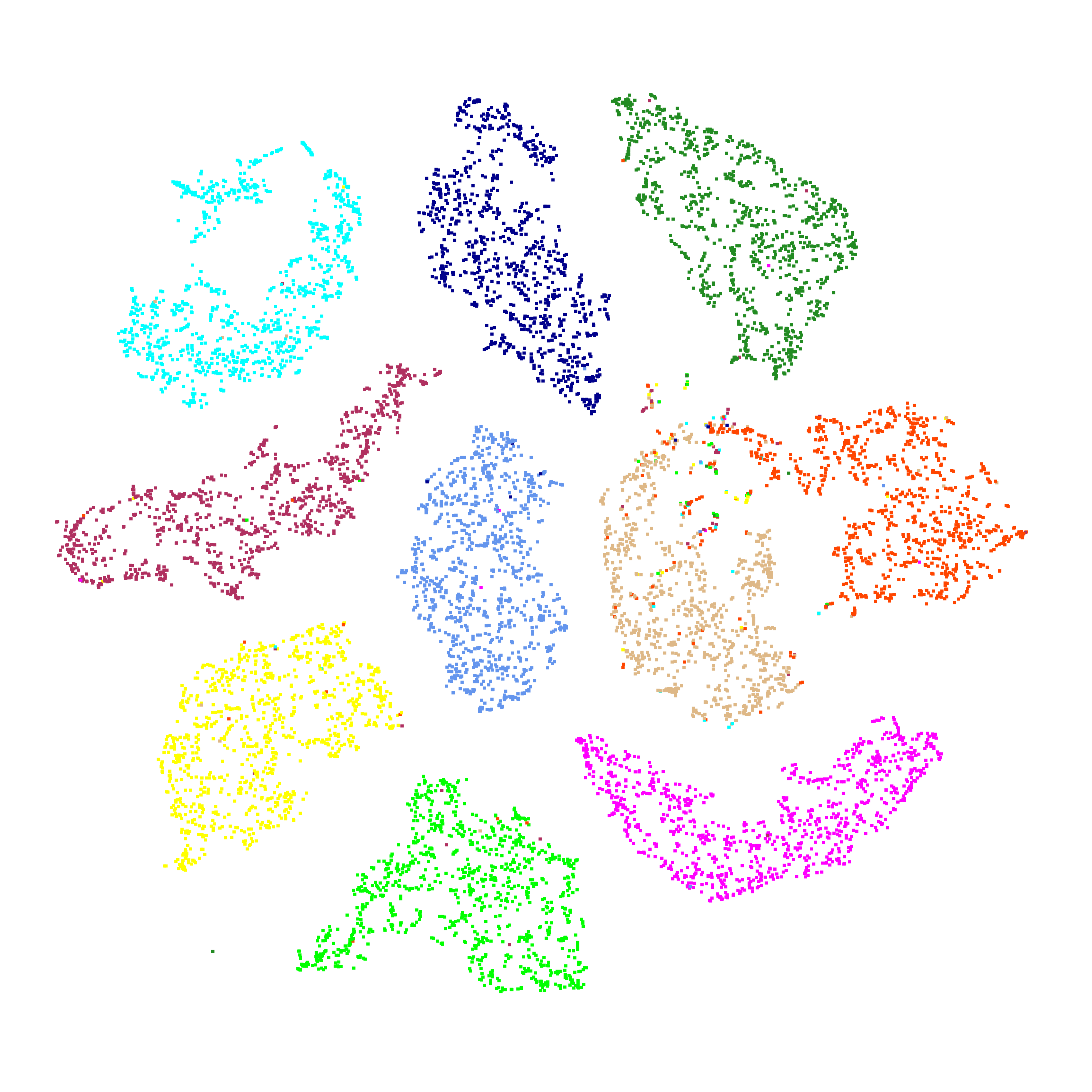}}
        \caption{MCL without SPA}
        \label{fig:tsne-mclwospa}
    \end{subfigure}
    \begin{subfigure}{.49\linewidth}
        \centering
        \frame{\includegraphics[width=.95\linewidth,height=.9\linewidth]{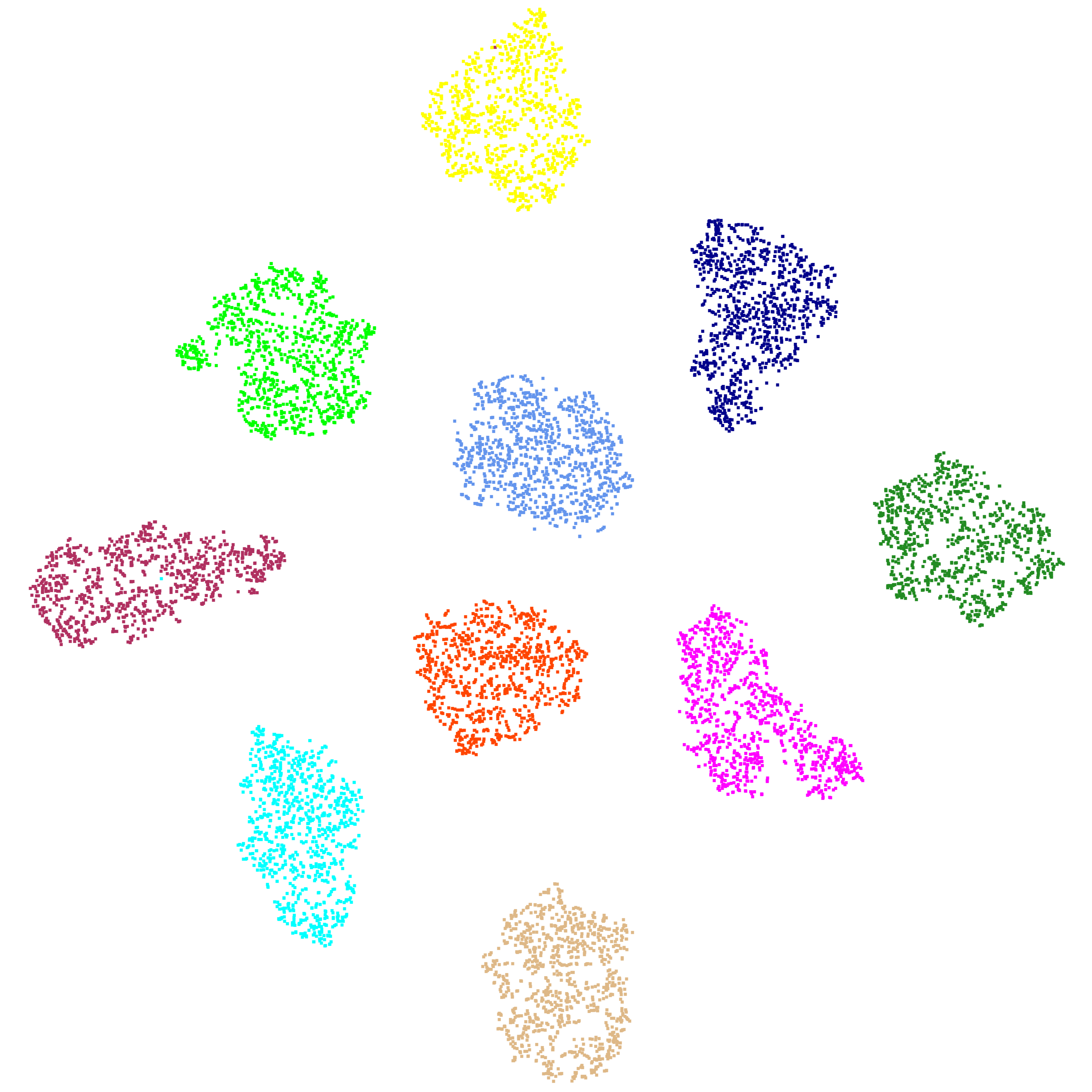}}
        \caption{MCL}
        \label{fig:tsne-mcl}
    \end{subfigure}
    \par \medskip
    \caption{$t$-SNE visualization of CIFAR-10 trained representation. Each color denotes 10 respective class labels in CIFAR-10 dataset. Stochastic Positive Attraction (SPA) indicates component in our model (MCL) and SupCLR is another \textit{task-specific} variant of SimCLR. While other models show blurry decision boundaries between some class pairs, MCL forms dense cluster for each class label while preserving their unique individual representation of respective data point in each cluster. See Section \ref{sec:MCL} for further details.}
    \label{fig:tsne}
\end{figure}

    Over the past few years, machine learning has achieved immense success surpassing human-level performance in many tasks, such as classification, segmentation, and object detection \cite{tan2019efficientnet,tan2020efficientdet,chen2020simple}.
    However, such a well-trained model assigns arbitrary high probability \cite{hein2019relu} on the unfamiliar test samples, since most machine learning systems generally depend on the closed-set assumption (\textit{i.e.}, i.i.d. assumption).
    This phenomenon may lead to a fatal accident in safety-critical applications like medical-diagnosis or autonomous driving.
    \textit{Anomaly detection}\footnote{also termed \textit{out-of-distribution detection}, \textit{novelty detection}, or \textit{outlier detection} in the contemporary machine learning context.} is a research area that aims to circumvent such symptoms by identifying whether the test samples come from in-distribution or not.
    A flurry of recent deep-learning based models, including reconstruction based \cite{oza2019c2ae,li2018anomaly}, density estimation based \cite{malinin2018predictive}, post-processing methods \cite{lee2018simple,liang2017enhancing}, and self-supervised learning methods \cite{golan2018deep,hendrycks2019using,tack2020csi,winkens2020contrastive}, have been proposed for the task and have shown noticeable progress.
    
    Among the numerous approaches mentioned, self-supervised learning (SSL) is in the limelight and validating its superiority over previous methods in various research areas \cite{devlin2018bert,chen2020simple}.
    Since it is unfeasible to access \textit{out-of-distribution} (OOD) data in most real-world scenarios, the ability of SSL to learn complex and diverse representations without additional labels is receiving much attention from anomaly detection lately.
    \cite{golan2018deep} is one of the earlier works to identify the potential of SSL and has proposed a simple yet effective technique that aims to learn intrinsic features within in-distribution (IND) samples via auxiliary tasks (\textit{e.g.}, predicting flip, rotation, or translation of input data).
    Furthermore, \cite{hendrycks2019using} confirmed that using such auxiliary tasks not only helps to determine anomalous samples but also helps to defend against adversarial attacks.
    More recent works \cite{tack2020csi,winkens2020contrastive} exploit contrastive learning (CL), especially SimCLR \cite{chen2020simple}, that learns individual data representations in a \textit{task-agnostic} way by maximizing the agreement between differently augmented views of the same image while repelling others in the batch.

    SimCLR obtains effective individual representation for each data point, as well as clustered representations for each class, even without any human label or supervision. (See Fig. \ref{fig:tsne-simclr}.)
    However, its \textit{task-agnostic} feature results in blurry boundaries between each cluster, so it requires a fine-tuning process used for some downstream tasks (\textit{e.g.}, multi-class classification).
    Such process undermines expression ability well-learned through SimCLR, given that most of the fine-tuning procedure leverages \textit{cross-entropy loss} and it solely considers class labels of the data without taking into account unique characteristics of the data or similarity between them.
    Consequently, the fine-tuned model often assigns high confidence probabilities to OOD input, reducing the distributional discrepancy between IND and OOD \cite{hein2019relu}.
    
    Our foremost insight is that forming dense clusters without fine-tuning while preserving individual representations by inheriting the advantages of SimCLR will shape a more meaningful visual representation contrary to the \textit{pre-train then tune} paradigm, thus contributing to the effective detection of anomalous data.
    To this end, we propose a \textit{task-specific} variant of contrastive learning called \textit{masked contrastive learning} (MCL), which can shape more clear boundaries between each class. (See Fig.\ref{fig:tsne-mcl})
    The core idea of MCL is to generate a mask that can adjust the repelling-ratio properly by considering class labels in the batch.
    Experimental results show that MCL is more befitted to anomaly detection then SimCLR or its other \textit{task-specific} variant (\textit{i.e.}, SupCLR), which still exhibits blurry decision boundaries (See Figure 1.\ref{fig:tsne-supclr}).
    
    Moreover, contrary to the previous belief that the auxiliary self-supervision task (\textit{e.g.}, predicting flip, rotation, or translation of input data) does not substantially improve label classification accuracy \cite{hendrycks2019using}, we observe that it is possible to considerably improve both IND and OOD performance with a proper inference method.
    To this end, we propose \textit{self-ensemble inference} (SEI) that fully exploits ability learned from simple auxiliary self-supervision task in the inference phase.
    SEI enhances model performance in all situations without losing generality and can be used in any classifier.
    By combining our models, we can outperform previous state-of-the-art methods.

    Our main contributions are summarized as below:
    \begin{itemize}
    	\item We propose a novel extension to contrastive learning dubbed \textit{masked contrastive learning} which can shape dense class-conditional clusters.
    	\item We also propose an inference method called \textit{self-ensemble inference}  that fully leverages ability learned from auxiliary self-supervision tasks in test time. \textit{Self-ensemble inference} can further boost both IND and OOD performance.
    	\item We validate our approaches on various image benchmark datasets, where we obtain significant performance gain over the previous state-of-the-art.
    \end{itemize}

    The source code for our model is available online.\footnote{https://github.com/HarveyCho/MCL}

\begin{figure*}[htbp!]
    \centering
    \includegraphics[width=1\textwidth]{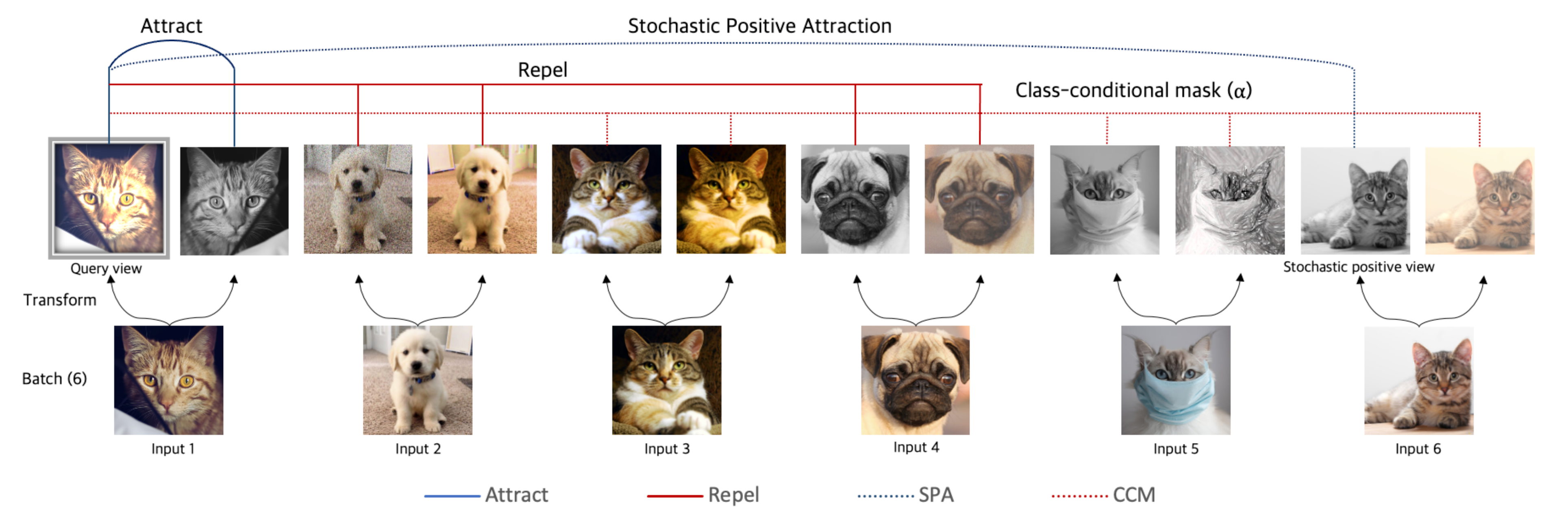}
    \caption{Description of MCL framework. Query view (gray-framed) attracts views connected with a blue-colored line and repels views connected with a red-colored line.}
    \label{fig:model}
\end{figure*}

\section{Masked Contrastive Learning}
\label{sec:MCL}

    As the name implies, our method adopts contrastive learning, particularly SimCLR, with two additional components: \textit{class-conditional mask} and \textit{stochastic positive attraction}; see Fig. \ref{fig:model}.
    In this section, we provide detailed explanations of each component in MCL. (See Section \ref{section:related-work} for further details regarding contrastive learning.)
    

    \subsection{Background: Contrastive Learning}

        Recent contrastive learning algorithms (\textit{e.g.}, SimCLR) learn representations by maximizing the agreement between differently augmented views of the same image while repelling others in the batch.
        Specifically, each image $\bm{x}_k$ from randomly sampled batch $\mathcal{B} = \{(\bm{x}_k, y_k)\}_{k=1}^N$
        is augmented twice, generating an independent pair of views $(\bar{\bm{x}}_{2k-1}, \bar{\bm{x}}_{2k})$ and augmented batch $\mathcal{\bar B}=\{(\bar{\bm{x}}_k, \bar y_k)\}_{k=1}^{2N}$, where labels of augmented views $\bar y_{2k-1}, \bar y_{2k}$ are equal to original label $y_{k}$.
        The augmented pair of views, $\bar{\bm{x}}_{2k-1} = t(\bm{x}_k)$ and $\bar{\bm{x}}_{2k} = t'(\bm{x}_k)$, are generated via independent transformation instance $t$ and $t'$, drawn from pre-defined augmentation function family $\mathcal{T}$.
        $(\bar{\bm{x}}_{2k-1}, \bar{\bm{x}}_{2k})$, then are passed sequentially through encoder network $f_{\theta}$ and projection head $g_{\phi}$, yielding latent vectors $(\bm{z}_{2k-1}, \bm{z}_{2k})$ that are utilized for the contrastive loss (\textit{i.e.}, NT-Xent):

        \begin{equation}
            \label{eq:nt-xent}
            \ell (i, j) =
                - \log
                \frac{
                    \text{exp} (
                        \text{sim}_{i, j}
                        /
                        \tau
                    )
                }{
                    \sum_{k = 1}^{2N}
                        \mathbb{1}_{[k \neq i]}
                        \text{exp} (
                            \text{sim}_{i, k}
                            /
                            \tau
                        )
                }
            ,
        \end{equation}

        \noindent where $\text{sim}_{i, j} = \bm{z}_i^\top \bm{z}_j / (\left\| \bm{z}_i \right\| \left\| \bm{z}_j \right\|)$ denotes cosine similarity between pair of latent vectors in $(\bm{z_i},\bm{z_j})$ and $\tau$ stands for temperature hyper-parameter.
        The final objective is to minimize Eq. \ref{eq:nt-xent} over positive pairs, which maps the input into effective individual representation in a \textit{task-agnostic} way:

        \begin{equation}
            \label{eq:loss-cl}
            \mathcal{L}_\text{SimCLR} =
                \frac{1}{2N}
                \sum_{k=1}^N
                    \left[
                        \ell (2k - 1, 2k)
                        +
                        \ell (2k, 2k - 1)
                    \right]
            .
        \end{equation}


    \subsection{Class-Conditional Mask}

        The benefit of CL in anomaly detection has been reported recently \cite{winkens2020contrastive,tack2020csi}.
        Nonetheless, we found that well-formed representations from CL, which facilitate distinguishing anomalous data, are lost during the fine-tuning procedure.
        Due to the \textit{task-agnostic} nature of CL, however, the fine-tuning steps are essential, making it difficult to avoid the aforementioned phenomenon.
        MCL mitigates such symptoms by injecting \textit{task-specific} characteristics to existing CL, resulting in fine-tuning procedure inessential.
        One key component in MCL is \textit{class-conditional mask} (CCM) which is a simple yet effective masking technique that adaptively determines the repelling-ratio considering the label information in each $\mathcal{\bar B}$.
        CCM can be defined as follows:

        \begin{equation}
            \label{eq:mask-function}
            \text{CCM}(i, j) =
                \left\{
                	\begin{array}{ll}
                		\alpha  & \mbox{if } \, \bar{y}_i = \bar{y}_j \\
                		1 / \tau     & \mbox{if } \, \bar{y}_i \neq \bar{y}_j,
                	\end{array}
                \right
        .
        \end{equation}
        
        \noindent where $0 < \alpha < 1 / \tau$.
        CCM alters the temperature for the same label views to a smaller value $\alpha$, so that query view repels views with the same label relatively small amount compared to views with different labels.
        The generated CCM is then multiplied to the similarity score in Eq. \ref{eq:nt-xent}, modifying the previous SimCLR loss to the following equation.
        \begin{equation}
            p_\text{ccm}(i, j) =
                \frac{
                    \text{exp} (
                        \text{sim}_{i, j}
                        /
                        \tau
                    )
                }{
                    \sum_{k = 1}^{2N}
                        \mathbb{1}_{[k \neq i]}
                        \text{exp} (
                            \text{sim}_{i, k}
                            \text{CCM}(i, k)
                        )
                }
            ,
        \end{equation}

        \begin{equation}
            \label{eq:nt-xent-ccm}
            \ell_\text{ccm} (i, j) =
                - \log
                p_\text{ccm}(i, j)
            ,
        \end{equation}

        \begin{equation}
            \label{eq:loss-ccm}
            \mathcal{L}_{\text{ccm}} =
                \frac{1}{2N}
                \sum_{k=1}^N
                    \left[
                        \ell_\text{ccm} (2k - 1, 2k)
                        +
                        \ell_\text{ccm} (2k, 2k - 1)
                    \right]
            .
        \end{equation}

        \noindent Penalizing a small ratio $\alpha$ to positive views restrains respective representation in the same cluster from being too similar to each other, making individual data representation more distinctive.


    \subsection{Stochastic Positive Attraction}
    \label{section:SPA}

        As can be seen in Fig. \ref{fig:tsne-mclwospa}, CCM promotes a more label-wise cluster compared to SimCLR.
        Even in CCM, however, the core operating principle is still identical to SimCLR in that it only attracts the view from the same image while it repels remaining views within the batch.
        Due to this repulsive nature, each data representation gets more distant as training continues, and it leads to the formation of unsatisfactory scattered clusters.
        To alleviate this phenomenon, we add another component named \textit{stochastic positive attraction} (SPA), an additional attraction with the stochastically sampled view in the batch.
        Specifically, SPA attracts query $\bar{\bm{x}}_i$ with stochastic positive sample $(\bar{\bm{x}}_{j},\bar y_{j}) \sim \mathcal{U}(\mathcal{\bar B}^{+}_{i})$ in the positive augmented batch $\mathcal{\bar B}^{+}_i$ for query $\bar{\bm{x}}_i$, where $\mathcal{U}$ refers to the discrete uniform distribution.
        The positive augmented batch $\mathcal{\bar B}^{+}_{i}$ for query $\bar{\bm{x}}_i$ contains views with the same label except views from its parent image $\bm{x}_{(i - 1) \setminus 2}$, where the symbol $\setminus$ denotes integer quotient operator:

        \begin{equation}
            \mathcal{\bar B}^{+}_{i} = \{
                (\bar{\bm{x}}_k, \bar y_k) \in \mathcal{\bar B}
                \mid
                \bar{y}_k = \bar{y}_i
                \text{ and }
                (k - 1) \setminus 2 \neq (i - 1) \setminus 2
            \}.
        \end{equation}

        \noindent
        CCM is also used for negative views with the additional constraint which excludes views from its parent image.
        SPA for query view $\bar{x}_i$ now can be defined as follows:

        \begin{equation}
            \label{eq:nt-xent-spa-inside}
            \resizebox{1\linewidth}{!}{$
                \displaystyle
                p_\text{spa}(i, j) =
                    \frac{
                        \text{exp} (
                            \text{sim}_{i, j}
                            /
                            \tau
                        )
                    }{
                        \sum_{k = 1}^{2N}
                        \mathbb{1}_{[(k - 1) \setminus 2 \neq (i - 1) \setminus 2 ]}
                            \text{exp} (
                                \text{sim}_{i, k}
                                \text{CCM}(i, k)
                            )
                    }
                ,
            $}
        \end{equation}

        \begin{equation}
            \label{eq:nt-xent-spa}
            \ell_\text{spa}(i) =
                \text{E}_{(\bar{\bm{x}}_{j}, \bar y_{j}) \sim \mathcal{U}(\mathcal{\bar B}^{+}_i)}
                \left[
                    - \log p_\text{spa}(i, j)
                \right]
            .
        \end{equation}

        \noindent
        The complete version of MCL is acquired by combining CCM and SPA, where the overall loss term being as follows:

        \begin{equation}
            \label{eq:loss-mcl}
            \mathcal{L}_\text{MCL} =
                \mathcal{L}_{\text{ccm}}
                +
                \frac{\lambda}{2N}
                \sum_{k=1}^{2N}
                    \ell_\text{spa} (k)
            ,
        \end{equation}

        \noindent where $\lambda$ denotes weight hyper-parameter for SPA loss.
        As a side note, $\alpha$ should meet certain conditions to function SPA correctly.
        Appendix A elaborates the above-mentioned conditions.
        

    \subsection{Training Auxiliary task in MCL}

        Training simple auxiliary self-supervision task along with the main downstream task is possible in MCL by adding constraint in CCM.
        Let $T_\text{main}$ be the main task with $C_\text{main}$ number of classes, $T_\text{aux}$ be an auxiliary task with $C_\text{aux}$ number of classes and corresponding augmented batch be $\mathcal{\bar B} = \{(\bar{\bm{x}}_i, \bar y_i^\text{main}, \bar y_i^\text{aux}) \}_{i=1}^{2N}$ with additional auxiliary task label.
        Then CCM with auxiliary self-supervision task can be defined as follows:
        \begin{equation}
            \label{eq:mask-function-aux-2}
            \text{CCM}_\text{aux}(i, j) =
                \left\{
                	\begin{array}{ll}
                		\alpha & \mbox{if } \, \bar{y}_i^\text{main} = \bar{y}_j^\text{main} \,\, \text{and} \,\, \bar{y}_i^\text{aux} = \bar{y}_j^\text{aux} \\
                		\beta  & \mbox{if } \, \bar{y}_i^\text{main} = \bar{y}_j^\text{main} \,\, \text{and} \,\, \bar{y}_i^\text{aux} \neq \bar{y}_j^\text{aux} \\
                		1 / \tau      & \mbox{otherwise.}
                	\end{array}
                \right
            .
        \end{equation}
        By simply setting $\beta=1/\tau$, it is possible to  train $C_\text{main} \times C_\text{aux}$ distinctive clusters for each ($\bar{y}_j^\text{main}$, $\bar{y}_j^\text{aux}$) pairs.
        Since the auxiliary task plays a complementary role to the main task, it is more plausible to form grouped clusters for respective main labels and to have distinctive clusters for each auxiliary label inside them.
        With appropriate constraint (\textit{i.e.}, $0 < \alpha < \beta < 1/ \tau$), MCL forms hierarchical clusters by dint of CCM.
        

\section{Inference}


    \subsection{Scoring Function in MCL}
    \label{section:scoring}

        Since there is no task-specific final layer in MCL, classification or anomaly detection are conducted via class-wise density estimation analogous to \cite{lee2018simple}, utilizing negative Mahalanobis distance $-d_{M}$ as a scoring function $s$:

        \begin{equation}
            \label{eq:mahalanobis-score}
            s_i(\bm{z}) = -d_M(
                \bm{z},
                \mu_i;
                \Sigma_i
            ) =
            (\bm{z} - \mu_i)^\top
            \Sigma^{-1}_i
            (\bm{z} - \mu_i)
            ,
        \end{equation}

        \begin{equation}
            S(\bm{x}) = \left[
                s_1(\bm{x}),
                s_2(\bm{x}),
                \cdots,
                s_{C_\text{main}}(\bm{x})
            \right],
        \end{equation}

        \noindent where $\bm{z} = g_\phi(f_\theta(\bm{x}))$, and $\mu_i$, $\Sigma_i$ refer to mean and covariance matrix of $n$-dimensional multivariate normal distribution (MND) $\mathcal{N}(\mu_i, \Sigma_i)$ for class $i \in I = \{1,2,\cdots, C_\text{main}\}$.
        Note that calculating MNDs for each class is a one-time operation acquired from training data.
        The vector $S(\bm{x})$ contains scores of each label for image $\bm{x}$ and the class label with highest score $i^* = \text{argmax}_{n \in I} S_n(\bm{x})$ is selected as a predictive label, where $S_{n}(\bm{x})$ denotes $n$-th element of $S(\bm{x})$.
        The corresponding IND score for predictive label $s_{i^*}(\bm{x})$ measures the confidence for predictive label $i^*$ which are used to distinguish OOD data, following the binary decision function $h_\delta$ from below:

        \begin{equation}
            \label{eq:score-based-detection}
            h_\delta(\bm{x}) =
                \left\{
                	\begin{array}{ll}
                		\text{IND} &  S_{i^*}(\bm{x}) \ge \delta \\
                		\text{OOD} &  S_{i^*}(\bm{x}) < \delta,
                	\end{array}
                \right
        .
        \end{equation}
        where $\delta$ denotes anomaly threshold.


    \subsection{Self Ensemble Inference}

        The key idea of SEI is to exploit the model's ability to discriminate within IND, learned through an auxiliary self-supervision task, in the inference phase.
        For example, consider predicting 4-directional rotations (from 0$^\circ$, 90$^\circ$, 180$^\circ$, to 270$^\circ$) is employed for an auxiliary task.
        Then SEI ensembles the results from corresponding the 4-rotated test images and derives calibrated index $i^*$ and corresponding score $s_{i^*}$.
        Specifically, let $i \in I = \{1, 2, ..., C_\text{main}\}$ be the main task label, and $j \in J = \{1, 2, ..., C_\text{aux}\}$ be the auxiliary task label.
        Then, $C_\text{main} \times C_\text{aux}$ number of MNDs  $\mathcal{N}(\mu_{i}^{(j)}, \Sigma_{i}^{(j)})$ are calculated for every label combinations.
        The test image $\bm{x}$ is augmented $C_\text{aux}$ times, yielding $\{\bm{x}^{(j)}\}_{j=1}^{C_\text{aux}}$.
        Each augmented test image $\bm{x}^{(m)}$ with the class label $y^\text{aux} = m$ is fed into the corresponding MND $\mathcal{N}(\mu_{i}^{(m)}, \Sigma_{i}^{(m)})$, where $i \in I$ and $j=m$, yielding a score vector $S^{(m)}(\bm{x})$:

        \begin{equation}
            s_{i}^{(j)}(\bm{x}) = -d_M(
                g_\phi(f_\theta(\bm{x})),
                \mu_{i}^{(j)};
                \Sigma_{i}^{(j)}
            ),
        \end{equation}

        \begin{equation}
            S^{(m)}(\bm{x}) = \left[
                s_{1}^{(m)}(\bm{x}),
                s_{2}^{(m)}(\bm{x}),
                \cdots,
                s_{C_\text{main}}^{(m)}(\bm{x})
            \right]
            .
        \end{equation}

        \noindent
        Our goal is to aggregate $\{S^{(j)}(\bm{x})\}_{j=1}^{C_\text{aux}}$ properly to make model more robust and reliable.
        We considered 3 different aggregation methods to extract predictive label $i^*$.
        The foremost intuitive way is averaging the main label scores across $\{S^{(j)}(\bm{x})\}_{j=1}^{C_\text{aux}}$.

        \begin{equation}
            i^*_\text{avg}(\bm{x}) =
                \underset{i \in I}{\text{argmax}}
                \frac{1}{|J|}
                \sum_{m\in J}
                    S_{i}^{(m)}(\bm{x})
            .
        \end{equation}

        \noindent
        Another variation is to select label index from the highest IND score:
        \begin{equation}
            i^*_\text{max}(\bm{x}) =
                \underset{i \in I}{\text{argmax}} \,
                \left\{
                \underset{m \in J}{\text{max}} \,
                    S_{i}^{(m)}(\bm{x})
                \right\}
            .
        \end{equation}

        \noindent
        The last aggregation is the weighted-average, which gives adaptive weights to each score in $S$.
        Weights for each score are computed per $j$ using the harmonic mean to penalize exceptionally low scores for better calibration:

        \begin{equation}
            W^{(m)}(\bm{x}) = \left(
                \sum_{n \in I}
                    \frac{1}{S_{n}^{(m)}(\bm{x})}
            \right)^{-1}
            .
        \end{equation}
        \begin{equation}
            i^*_\text{w-avg}(\bm{x}) =
                \underset{i \in I}{\text{argmax}}
                \frac{
                    \sum_{m \in J}
                        W^{(m)}(\bm{x})
                        S_{i}^{(m)}(\bm{x})
                }{
                    \sum_{m \in J}
                        W^{(m)}(\bm{x})
                }.
        \end{equation}
        Corresponding score to predictive label $i^*$ is used to distinguish OOD data following Eq. \ref{eq:score-based-detection}.
        Depending on the aggregation method, the effect that the model can achieve varies. Further details are elaborated in Section \ref{section:ablation} with experimental results.


\begin{table*}
    \setlength{\tabcolsep}{10pt}
    \begin{tabular}{l|c|ccccc}

        \toprule

            & & \multicolumn{4}{c}{AUROC} \\

            Training Method & Test Acc. & SVHN & LSUN(F) & ImageNet(F) & CIFAR-100 \\

        \midrule

            Baseline \cite{hendrycks2016baseline}
            & 93.6 &	89.9 & 84.3 &	88.0 &	86.4 \\

            ODIN$^\star$ \cite{liang2017enhancing}
            & 93.6	&	85.8	&	83.2 &	87.7 &	85.8 \\

            Mahalanobis Distance$^\star$ \cite{lee2018simple}
            & 94.1 &	99.1 &	87.9 &	90.9 & 88.2 \\

            Auxiliary Rotation \cite{hendrycks2019using}
            & 94.3 & 97.3 & 94.5 & 94.7 & 90.7\\

            Outlier Exposure$^\dagger$ \cite{hendrycks2018deep}
            & - & 98.4  & - & - & 93.3 \\

            SupCLR$^\ddagger$ \cite{khosla2020supervised}
            & 93.8 & 97.3 & 91.6 & 90.5 & 88.6 \\

            CSI \cite{tack2020csi} + SupCLR$^\ddagger$
            & 94.8 & 96.5  & 92.1 & 92.4 & 90.5 \\

            CSI + SupCLR + Ensemble$^\ddagger$
            & 96.1 & 97.9  & 93.5 & 94.0 & 92.2 \\

        \midrule

            Auxiliary Rotation + SEI
            & 95.8 & 98.4 & 95.7 & 95.8 & 92.3 \\

            MCL + SEI (ours) $^\ddagger$
            & 95.9 & 98.9  & 96.0 & 95.9 & 93.1 \\

            \textbf{MCL + SEI (ours)}
            & \textbf{96.4} & \textbf{99.3}  & \textbf{96.3} & \textbf{96.5} & \textbf{94.0} \\

        \bottomrule

    \end{tabular}

    \vspace{0.1cm}

    \begin{tabular}{ll}
        $^\star$ refers models with additional post-proccesing procedure. & $^\dagger$ denotes models with additional supervised (OOD) data.\\
    \end{tabular}
    \begin{tabular}{l}
        $^\ddagger$ denotes models that use ResNet-18 as a backbone network, and models without the mark use ResNet-34.
    \end{tabular}

    \vspace{-0.2cm}
    \caption{Test accuracy of in-domain classification and AUROC of OOD data for each model trained on CIFAR-10 dataset.}
    \label{tab:result-main}
\end{table*}

\begin{table*}
    \centering
    \setlength{\tabcolsep}{14pt}
    \begin{tabular}{l|c|ccccc}
        \toprule
            & & \multicolumn{4}{c}{AUROC} \\\
            Training Method & Test Acc. & SVHN & LSUN(F) & ImageNet(F) & CIFAR-100 \\
        \midrule

            SupCLR \cite{khosla2020supervised}
            & 93.8 & 97.3 & 91.6 & 90.5 & 88.6 \\

            MCL (Ours)
            & 93.1 & 97.9  & 93.8 & 93.6 & 90.8 \\

            MCL + SEI (Ours)
            & 95.9 & 98.9  & 96.0 & 95.9 & 93.1 \\

        \bottomrule

    \end{tabular}

    \caption{Performance comparison between MCL and SupCLR. Models are trained on CIFAR-10 dataset with ResNet-18.}
    \label{tab:result-main-cifar100}
\end{table*}

\section{Experiment}

    In this section, we demonstrate the effectiveness of MCL and SEI on several multi-class image classification datasets.

    \noindent
    \paragraph{Experiment configurations.} In the following experiments, we adopt ResNet-34 \cite{he2016deep} with a single projection head, following structure used to train CIFAR-10 in SimCLR.
    We also fixed hyper-parameters related to contrastive learning following SimCLR, which include transformation $\mathcal{T}= \{ $color jittering, horizontal flip, grayscale, inception crop$\}$, the strength of color distortion to 0.5, batch size to 1024, and temperature $\tau$ to 0.2, to keep our experiment tractable.
    For MCL hyper-parameters, we set $\alpha$ to 0.05, $\beta$ to 2.5, and $\lambda$ to 1 which meets  certain condition for MCL (See Appendix A for details).
    Unlike SimCLR, we used SGD optimizer with learning rate 1.2 (0.3 $\times$ batch size / 256), decay 1e-6, and momentum 0.9.
    Furthermore, we use a cosine annealing scheduler without any warm-up.

    \noindent
    \paragraph{Evaluation metrics.} To evaluate IND detection performance, we measured the label classification accuracy. For OOD detection performance, we used the \textit{area under the receiver operating characteristic curve} (AUROC), which is a threshold ($\delta$ in Section \ref{section:scoring}) free metric and the most common metric in anomaly detection literature. 
    Results with additional metrics (\textit{i.e.}, FPR@95, AUPR) can be found in Appendix C.

    \noindent
    \paragraph{Selecting self-supervision tasks.}
    Since the complex auxiliary task is not our primary concern, we considered rotation, horizontal flip, and translation as candidates for auxiliary tasks, which are simple and commonly used in the area of anomaly detection \cite{golan2018deep}.
    However, MCL contains inception crop \cite{szegedy2015going} and horizontal flip in contrastive transformation $\mathcal{T}$, so using translations or horizontal flip as an auxiliary task only confuses the model.
    To this end, we employed predicting 4-directional rotations (0$^{\circ}$, 90$^{\circ}$, 180$^{\circ}$, and 270$^{\circ}$) as our auxiliary task.


    \subsection{Multi-Class Anomaly Detection}

        We trained our model on CIFAR-10 \cite{krizhevsky2009learning} as IND, and used CIFAR-100, SVHN \cite{netzer2011reading}, ImageNet \cite{deng2009imagenet}, and LSUN \cite{yu2015lsun} datasets for OOD.
        Note that all the classes in OOD datasets are disjoint with CIFAR-10.
        In particular, for ImageNet and LSUN dataset, we use ImageNet-Fix and LSUN-Fix datasets \cite{tack2020csi} which are fixed versions of the previously released dataset \cite{liang2017enhancing}. 
        The previous version contains unintended artifacts caused by resizing large images.
        Appendix C provides detailed explanations regarding datasets with a visual explanation.
        
        \subsubsection{Main Results} 
        We compared MCL's performance with several other methods in OOD detection.
        Table \ref{tab:result-main} summarizes our experimental results.
        MCL shows significant performance gain over previous methods in all OOD datasets; moreover, it outperforms supervised method \cite{hendrycks2018deep}, which utilizes additional explicit OOD data. 
        
        \subsubsection{Comparison with SupCLR} 
        We compared MCL's performance with SupCLR \cite{khosla2020supervised}, another \textit{task-specific} variant of CL.
        Table \ref{tab:result-main-cifar100} summarizes performance comparison with SupCLR.
        In terms of test accuracy, SupCLR performs slightly better than MCL, while in terms of AUROC, MCL shows better performance.
        The superiority of MCL in detecting anomalies comes from several differences between the two frameworks.
        Specifically, SupCLR additionally attracts all same label views from $\mathcal{\bar B}_{i}^{+}$ in the augmented batch $\mathcal{\bar B}$.
        Since SupCLR forces all positive views to have a high similarity to the query view, the ability to distinguish data with the same label diminishes.
        Unlike SupCLR, MCL penalizes a small amount $\alpha$ to positive views, which endows an ability to discern each data with same label.
        (A more detailed comparison with MCL and explanation of SupCLR is in Appendix B.)


\begin{table}
    \centering
    \setlength{\tabcolsep}{13pt}
    \begin{tabular}{l|c|c}

        \toprule

            Model
            & Acc & AUROC \\

        \midrule

            Baseline (w/o NT-Xent)
            & 93.61 & 86.40 \\

            SimCLR (CE fine-tuned)
            & 93.88 & 87.56 \\
            
            SimCLR (joint fine-tuned)
            & 93.91 & 88.51 \\

        \midrule
            MCL (w/o SPA, w/o aux)
            & 91.41 & 85.03 \\

            MCL (w/o aux)
            & \textbf{94.35} & 89.49 \\

            MCL (CE fine-tuned)
            & 94.22 & 88.89 \\

            \textbf{MCL}
            & 94.03 & \textbf{91.12} \\


        \bottomrule

    \end{tabular}
    \caption{Ablation studies for each component in MCL.}
    \label{tab:result-ablation-overall}
\end{table}

\begin{table}
    \centering
    \setlength{\tabcolsep}{10pt}
    \begin{tabular}{l|c|c|c}

        \toprule

            Model & Agg & Acc & AUROC \\

        \midrule

            \multirow{1}{*}{MCL + w/o SEI}

            & -
            & 94.03 & 91.12 \\

        \midrule

            \multirow{3}{*}{MCL + 4-way SEI}

            & avg
            & 94.68 & 93.20 \\

            & max
            & 95.97 & 92.30 \\

            & w-avg
            & 96.12 & 93.29 \\

        \midrule

            \multirow{3}{*}{\textbf{MCL + 8-way SEI}}

            & avg
            & 94.74 & 93.37 \\

            & max
            & 96.40 & 92.00 \\

            & \textbf{w-avg}
            & \textbf{96.43} & \textbf{94.06} \\

        \bottomrule

    \end{tabular}
    \caption{Ablation studies for SEI. MCL is trained with additional 4-way rotations auxiliary task.}
    \label{tab:result-ablation-sei}
\end{table}

    \subsection{Ablation Study}
    \label{section:ablation}

        In this section, we perform an ablation study on our proposed methods, along with baselines.
        In all experiments, we treated CIFAR-10 as IND and CIFAR-100 as OOD.

        \subsubsection{Masked Contrastive Learning}
        \label{subsection:MCL}
            We conduct ablation experiments to explore the effectiveness of the main components (CCM, SPA, and auxiliary 4-way rotation task) in MCL. 
            Tab. \ref{tab:result-ablation-overall} reports our ablation experiments along with baseline models.
            For SimCLR based models, we fine-tuned the pre-trained model in two ways.
            One way is to use \textit{cross-entropy loss} which is a traditional fine-tuning method in the classification task, and the other way is to use \textit{cross-entropy loss} along with SimCLR loss (Eq.\ref{eq:loss-cl}) jointly \cite{winkens2020contrastive}.
            The two methodologies show almost the same accuracy, while there is a slight difference in AUROC.
            Our justification for this phenomenon is due to the nature of the \textit{cross-entropy loss}, which solely reflects the class label leading diminish in the distributional discrepancy between IND and OOD.
            We conjectured that the additional SimCLR loss in joint fine-tuning mitigates this phenomenon and shows better AUROC performance.
            This phenomenon is more evident when applied to MCL, showing substantial performance degradation in AUROC.
            Therefore, we used MCL with neither a fine-tuning procedure nor any \textit{task-specific} layer on the top level of the network.

        \subsubsection{Self-Ensemble Inference 1}
    
\begin{figure}[t!]
    \centering
    \includegraphics[width=0.94\linewidth]{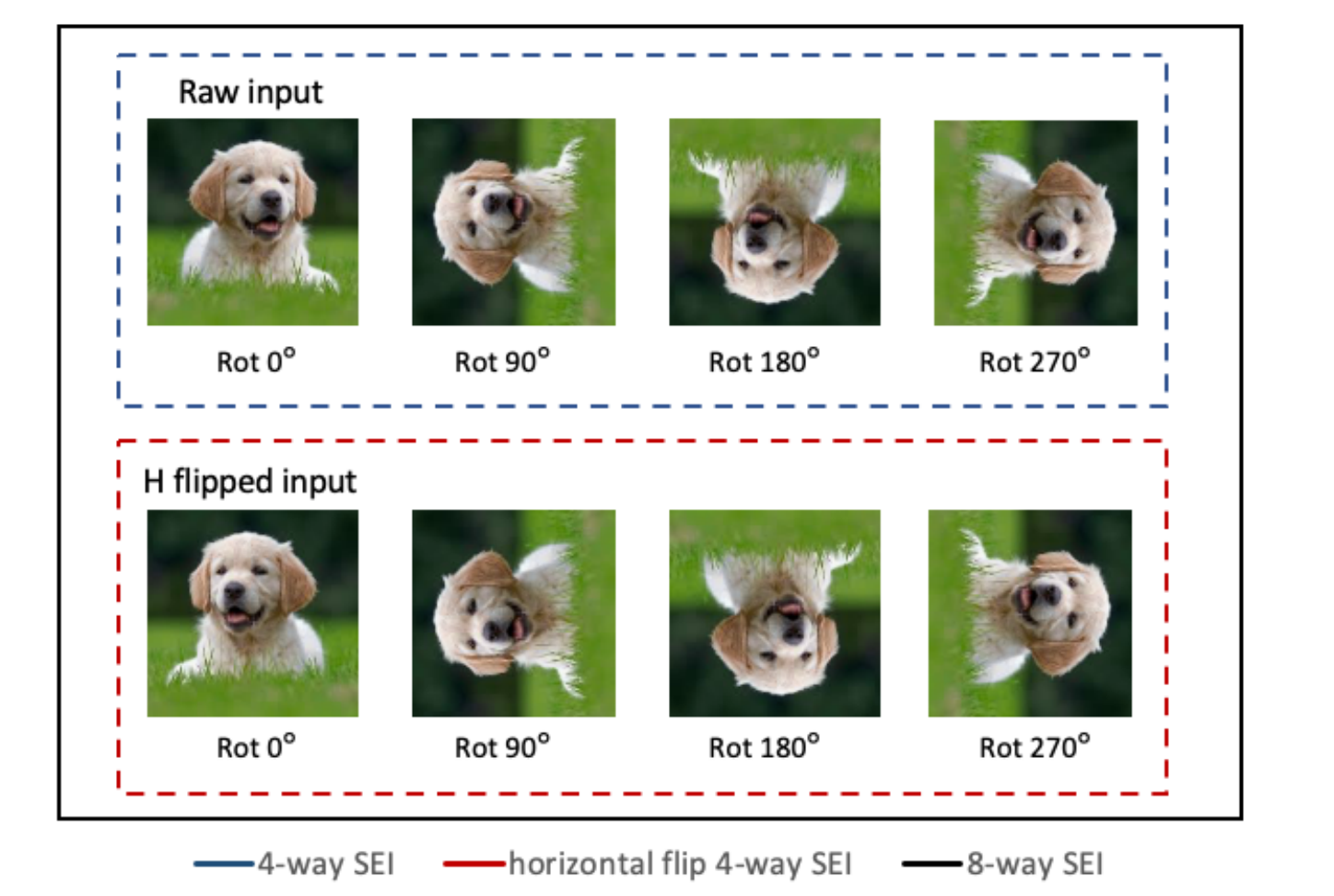}
    \caption{4-way SEI and 8-way SEI.}
    \label{fig:8-waySEI}
\end{figure}

            In this part, we share our findings on SEI with its variations (average, maximum, and weighted-average).
            Since we employed a 4-directional rotation prediction as our auxiliary task, SEI is done in a 4-way correspondingly.
            It is also possible to add additional 4-way SEI with a horizontally flipped image, which we dubbed 8-way SEI. (Fig. \ref{fig:8-waySEI} provides a visual explanation.)
            8-way SEI follows the same strategy introduced earlier. 
            The only difference is the number of augmented images to ensemble. 
            As claimed in \cite{hendrycks2019using}, learning the auxiliary task alone can not improve accuracy.
            But with the help of SEI, we were able to achieve performance gains in both accuracy and AUROC regardless of its variations as can be seen in Tab. \ref{tab:result-ablation-sei}.
            Interestingly, the effect of each ensemble differs depending on the aggregation method.
            For example, average SEI makes the model robust to input variation which is a commonly known benefit of the ensemble, bringing noticeable gain in AUROC contrary to accuracy.
            On the other hand, maximum SEI yields a significant gain in accuracy, which indicates MCL's prediction with high confidence score is quite precise.
            The weighted-average SEI absorbs the advantages of both ensembles by assigning adaptive weights to its score.
            As a side note, the SEI can be used for the general classifier trained with an auxiliary task, which also yields significant performance gains as can be seen in Tab. \ref{tab:result-main}.

\begin{table}
    \centering
    \setlength{\tabcolsep}{10pt}
    \begin{tabular}{l|c|c|c}

        \toprule

            Model & Agg & Acc & AUROC \\

        \midrule

            \multirow{2}{*}{MCL + w/o Aux}

            & -
            & 94.35 & 90.49 \\
            
            
            & w-avg
            & 83.96 & 71.17 \\

        \midrule

            \multirow{2}{*}{MCL + DA }

            & -
            & 92.55 & 90.07 \\
            & w-avg
            & 94.70 & 92.08 \\

        \midrule

            \multirow{2}{*}{\textbf{MCL + Aux }}

            & -
            & 94.03 & 91.12 \\
            
            
            & \textbf{w-avg}
            & \textbf{96.43} & \textbf{94.06} \\

        \bottomrule

    \end{tabular}
    \caption{Ablation studies for SEI. data augmentation is abbreviated to DA. Symbol - indicates a model without SEI.}
    \label{tab:appendix-result-ablation-sei}
\end{table}

\begin{figure*}[tp!]
    \begin{subfigure}{.49\linewidth}
        \centering
        \includegraphics[width=.95\linewidth]{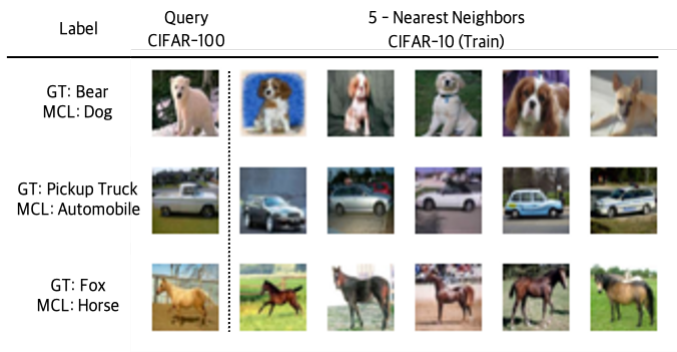}
        \caption{OOD samples with high IND scores.}
        \label{fig:case-ood}
    \end{subfigure}
    \begin{subfigure}{.49\linewidth}
        \centering
        \includegraphics[width=.95\linewidth]{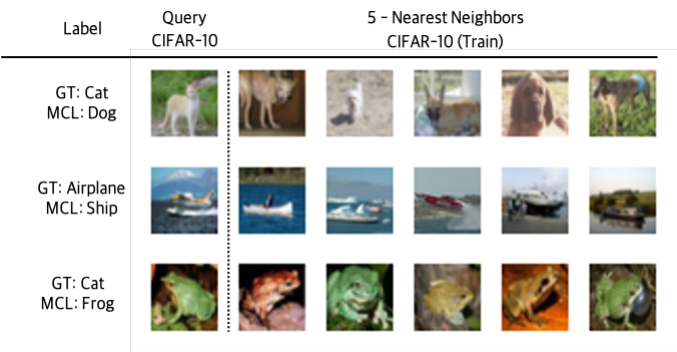}
        \caption{Wrongly classified samples from IND.}
        \label{fig:case-ind}
    \end{subfigure}
    \caption{Case studies on OOD samples with high confidence and wrongly classified IND samples.}
    \label{fig:case}
\end{figure*}

\begin{figure}[t!]
    \includegraphics[width=\linewidth]{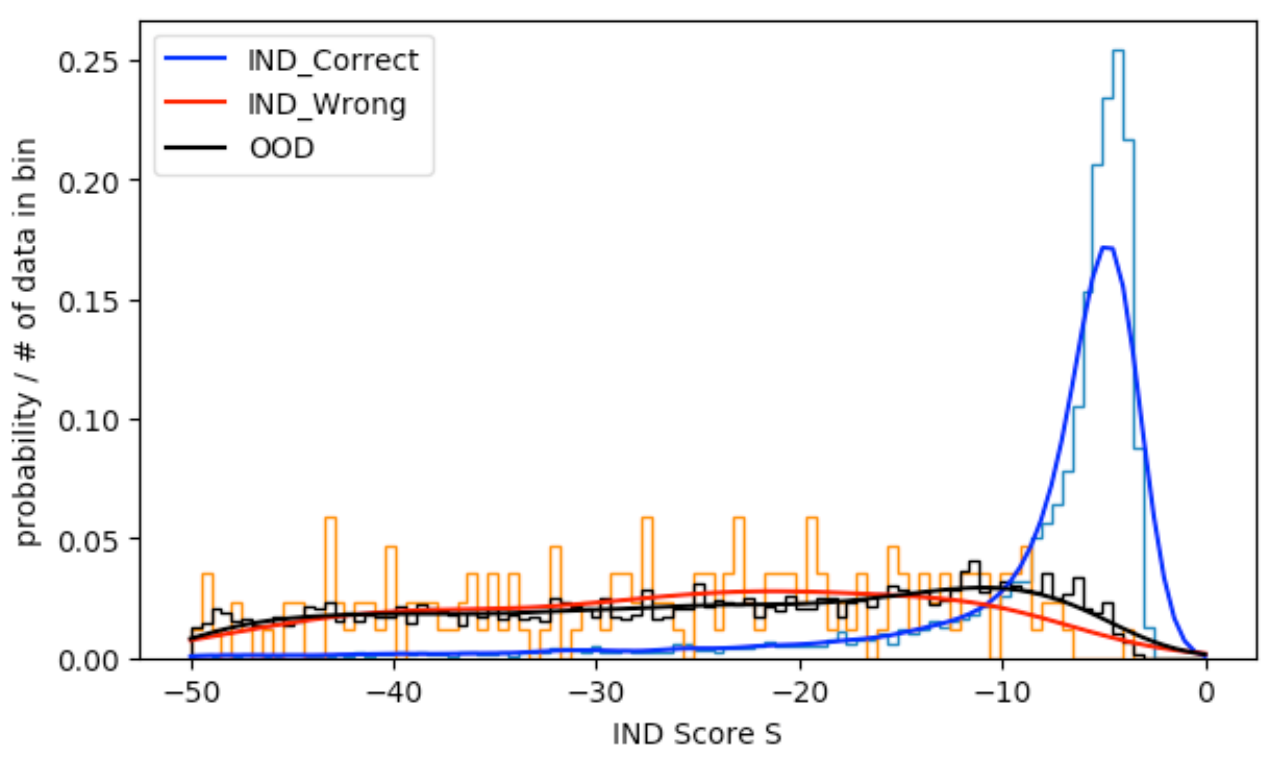}
    \caption{Histogram and PDF for correctly classified IND, wrongly classified IND and OOD distribution.}
    \label{fig:distribution}
\end{figure}

        \subsubsection{Self-Ensemble Inference 2}
            We also conduct ablation experiments on SEI and the relationship with the auxiliary task.
            We applied SEI to 3 differently trained model as follows:
            
            \begin{itemize}
                \item \textbf{MCL without auxiliary task:} MCL is trained neither with auxiliary task nor data augmentation
                \item \textbf{MCL with data augmentation:} MCL trained with additional 4-way rotated images without rotation labels.
                \item \textbf{MCL with auxiliary task:} MCL is trained with additional 4-way rotated images with rotation labels.
            \end{itemize}
            
            Tab. \ref{tab:appendix-result-ablation-sei} summarizes our extended ablation study for SEI (Tab. \ref{tab:appendix-result-ablation-overall} shows the performance of additional metrics).
            Applying SEI to MCL without auxiliary task only confuses the model and undermines both IND and OOD performance substantially, as rotated images are unseen during training phase.
            If the rotated image is augmented without a label, the model performance is degraded but the SEI shows a slight improvement.
            Finally, when the auxiliary task is explicitly trained with the main downstream task, SEI shows significant gain without any performance degradation.
            Such experimental results reveal that it is necessary to additionally train the auxiliary task to use SEI properly.

    \subsection{Qualitative results}

        In this section, we analyze the data distribution along with several failure cases of our model.
        As can be seen in Fig. \ref{fig:distribution}, both OOD data distribution and wrongly classified data distribution have lower scores compared to correctly classified samples, which indicates that our model can measure predictive uncertainty quite precisely.
        Furthermore, to analyze our failure cases in detail, we conducted a case study on OOD samples with high confidence and wrongly classified IND samples.
        In MCL, representation is learned based on semantic similarity, so it is feasible to conjecture model's decision-making process via observing nearest neighbors of the input.
        For most failure cases, their predicted label and ground-truth label (GT) belongs to the same super-class regardless of the aforementioned failure types.
        For example, MCL predicts a bear (OOD) as a dog (IND); likewise, a cat as a dog (both IND), which belongs to the same super-class, mammal.
        As a side note, few data were mislabeled, as can be seen in Fig. \ref{fig:case}.(a).(GT: Fox)
        Fig. \ref{fig:appendix-case} in the Appendix provides more case studies.


\section{Related Work}
\label{section:related-work}

    \noindent
    \paragraph{Anomaly detection.}
    Recent approaches in OOD can be categorized as follows:
    reconstruction based \cite{oza2019c2ae,li2018anomaly}, density estimation based \cite{malinin2018predictive}, post-processing based \cite{lee2018simple,liang2017enhancing}, and self-supervised learning based.
    Self-supervised learning based methods can be split again into auxiliary self-supervision based  \cite{golan2018deep,hendrycks2019using} and contrastive learning based \cite{tack2020csi,winkens2020contrastive}.
    Our method belongs to self-supervised learning, which exploits both auxiliary self-supervision task and contrastive learning.

    \noindent
    \paragraph{Contrastive learning.}
    Contrastive learning is a specific framework of self-supervised learning, which has shown impressive results in visual representation learning tasks \cite{chen2020simple,chen2020big}.
    Most recent work in OOD \cite{tack2020csi,winkens2020contrastive} report that employing CL improves OOD performance.
    Our work goes further from the previous papers and proposes a \textit{task-specific} variant of CL.
    \cite{khosla2020supervised} proposed SupCLR, another \textit{task-specific} variant of SimCLR, which is a noteworthy work.
    Similar to MCL, SupCLR also leverages label information in the batch while training and shows superior accuracy over SimCLR.
    Despite its performance in IND accuracy, representation from SupCLR which is not entirely appropriate for discerning anomalous data, as it attracts all same label views which discrepancy in each class disappears.
    We provide a comprehensive survey of previous works in Appendix D.


\section{Conclusion}

    In this paper, we propose a novel training method called \textit{masked contrastive learning} (MCL) and an inference method called \textit{self-ensemble inference} (SEI).
    MCL can shape class-conditional clusters by inheriting advantages of CL and  SEI fully leverages trained features from auxiliary self-supervised tasks in the inference phase.
    By combining our methods, our model reaches the new state-of-the-art performance.

\section*{Acknowledgements}
    We thank Taeuk Kim, Ye-seul Song and the anonymous reviewers for their thoughtful feedback and comments.
 

\bibliographystyle{style/named}
\bibliography{reference}


\cleardoublepage
\appendix
\noindent{\Large\bfseries Appendix}
\vspace{0.4cm}

\section{Efficiency of SPA}

    As we mentioned in the paper (Sec. 2.3), SPA possesses the potential to make respective clusters denser.
    However, such an ability of SPA comes when a certain condition is met.
    Otherwise, SPA either loses the ability to assemble scattered clusters, or forces every data points to converge near the centroid of each cluster, making the data points within the cluster indistinguishable.
    In the following section, we explore the two aforementioned conditions theoretically.

    \subsection{Attraction Condition}
        
        This section details the condition to maintain the ability of SPA to assemble a scattered cluster. Note that the role of SPA is to alleviate the same-class repulsion in MCL.
        
        Let $q$ be a query index in the augmented batch $\mathcal{\bar B}$ of size $2N$ and $r$ be the index of a stochastic positive sample from $\mathcal{\bar B}^{+}_{q}$.
        If the sign of the gradient of $\mathcal{L}_\text{spa} = \frac{1}{2N} \sum_{i=1}^{2N} \ell_\text{spa}(i)$ \textit{w.r.t.} $\text{sim}_{q, r}$ is negative, then the cosine similarity between two vectors $\bar{\bm{z}}_q, \bar{\bm{z}}_r$ gets higher as training continues.
        Such fact indicates that the ability of SPA comes when the gradient of SPA loss is negative.

        From SPA loss in Sec. 2.3, 
        
        \begin{equation}
            \resizebox{1\linewidth}{!}{$
                \displaystyle
                p_\text{spa}(i, j) =
                    \frac{
                        \text{exp} (
                            \text{sim}_{i, j}
                            /
                            \tau
                        )
                    }{
                        \sum_{k = 1}^{2N}
                        \mathbbm{1}_{[(k - 1) \setminus 2 \neq (i - 1) \setminus 2 ]}
                            \text{exp} (
                                \text{sim}_{i, k}
                                \text{CCM}(i, k)
                            )
                    }
                ,
            $}
        \end{equation}
        
        \begin{equation}
            \begin{split}
                \ell_\text{spa}(i)
                &=
                    \mathbb{E}_{(\bar{\bm{x}}_{j}, \bar y_{j}) \sim \mathcal{U}(\mathcal{\bar B}^{+}_i)}
                    \left[
                        - \log p_\text{spa}(i, j)
                    \right]
                \cr
                &\doteq
                    \frac{1}{|\mathcal{\bar B}^{+}_i|}
                    \sum_{(\bar{\bm{x}}_{j}, \bar y_{j}) \in \mathcal{\bar B}^{+}_i}
                        - \log p_\text{spa}(i, j)
                .
            \end{split}
        \end{equation}
        
        \noindent Then, gradient of $\mathcal{L}_\text{spa}$ \textit{w.r.t.} $\text{sim}_{q, r}$ can be calculated as follows: 
        
        \begin{equation}
            \begin{split}
                \frac{
                    \partial \mathcal{L}_{\text{spa}}
                }{
                    \partial \text{sim}_{q, r}
                }
                &=
                    \frac{1}{2N}
                    \sum_{i=1}^{2N}
                        \frac{
                            \partial \ell_\text{spa}(i)
                        }{
                            \partial \text{sim}_{q, r}
                        }
                    =
                        \frac{1}{2N}
                        \frac{
                            \partial \ell_\text{spa}(q)
                        }{
                            \partial \text{sim}_{q, r}
                        }
                \cr
                &=
                    \frac{1}{2N}
                    \frac{1}{|\mathcal{\bar B}^+_q|}
                    \left(
                        - \frac{1}{\tau}
                        + |\mathcal{\bar B}^+_q|
                          \alpha
                          p_\text{spa}(q, r)
                    \right)
                \cr
                &=
                    \frac{1}{2N}
                    \left(
                        \alpha p_\text{spa}(q, r)
                        - \frac{1}{\tau |\mathcal{\bar B}^+_q|}
                    \right)
                \cr
                &<
                    \frac{1}{2N}
                    \left(
                        \alpha
                        - \frac{1}{\tau |\mathcal{\bar B}^+_q|}
                    \right)
                ,
            \end{split}
        \end{equation}
        where $\text{CCM}(q, r) = \alpha$ and $p_\text{spa}(i, j) < 1$ for any $(i, j)$.

        To this end, $\alpha$ should be selected carefully which met following condition:
        \begin{equation}
            \label{eq:attraction-condition}
            \alpha < \frac{1}{\tau |\mathcal{\bar B}^+_q|}
            .
        \end{equation}


    \subsection{Convergence Condition}

        This section details the condition to restrain SPA from forcing every data point to converge near the centroid of each cluster.

        When  $\alpha$ is set to an extremely small value, the attraction from SPA becomes dominant compared to the repulsion of CCM, leading respective data points within each cluster to converges near its centroid.
        In concrete, let $\bar i$ be the index for another augmented view from same parent image $x_{(i-1)\setminus2}$.
        Then we can revise MCL loss as follows:
        
        \begin{equation}
            \begin{split}
                \mathcal{L}_{\text{MCL}}
                &=
                    \mathcal{L}_{\text{ccm}}
                    +
                    \lambda
                    \mathcal{L}_{\text{spa}}
                \cr
                &=
                    \frac{1}{2N}
                    \sum_{i=1}^{2N}
                        \ell_\text{ccm}(i, \bar i)
                    +
                    \lambda
                    \mathcal{L}_{\text{spa}}
                .
            \end{split}
        \end{equation}
    
        \noindent
        The overall gradient for MCL becomes:
    
        \begin{equation}
            \frac{
                \partial
                \mathcal{L}_{\text{MCL}}
            }{
                \partial
                \text{sim}_{q, r}
            }
            =
            \frac{1}{2N}
            \sum_{i=1}^{2N}
                \frac{
                    \partial \ell_\text{ccm}(i, \bar i)
                }{
                    \partial \text{sim}_{q, r}
                }
            +
            \frac{
                \partial \mathcal{L}_{\text{spa}}
            }{
                \partial \text{sim}_{q, r}
            }
            .
        \end{equation}
        
        \noindent
        By the definition of $\ell_{ccm}$ from Sec. 2, 
        
        \begin{equation}
            p_\text{ccm}(i, j) =
                \frac{
                    \text{exp} (
                        \text{sim}_{i, j}
                        /
                        \tau
                    )
                }{
                    \sum_{k = 1}^{2N}
                        \mathbbm{1}_{[k \neq i]}
                        \text{exp} (
                            \text{sim}_{i, k}
                            \text{CCM}(i, k)
                        )
                }
            ,
        \end{equation}
        
        \begin{equation}
            \ell_\text{ccm} (i, j) =
                - \log
                p_\text{ccm}(i, j)
            ,
        \end{equation}
        where $\mathcal{\bar B}^{+}_{q}$ guarantees $r \neq \bar q$.
        Then, the gradient of $\partial \ell_\text{ccm}(i, \bar i)$  \textit{w.r.t.} $\text{sim}_{q, r}$ becomes:
        
        \begin{equation}
            \frac{1}{2N}
            \sum_{i=1}^{2N}
                \frac{
                    \partial \ell_\text{ccm}(i, \bar i)
                }{
                    \partial \text{sim}_{q, r}
                }
            =
            \frac{1}{2N}
            \frac{
                \partial \ell_\text{ccm}(q, \bar q)
            }{
                \partial \text{sim}_{q, r}
            }
            =
            \frac{1}{2N}
            \alpha
            p_\text{ccm}(q, r)
            .
        \end{equation}
        
        By Eq. 9 and Eq. 3, the respective term of MCL can be rewritten as follows:
        
        \begin{equation}
            \begin{split}
                \frac{
                    \partial
                    \mathcal{L}_{\text{MCL}}
                }{
                    \partial
                    \text{sim}_{q, r}
                }
                &=
                    \frac{1}{2N}
                    \alpha p_\text{ccm}(q, r)
                    +
                    \frac{\lambda}{2N}
                    \left(
                        \alpha p_\text{spa}(q, r)
                        - \frac{1}{\tau |\mathcal{\bar B}^+_q|}
                    \right)
                \cr
                &<
                    \frac{1}{2N}
                    \alpha
                    +
                    \frac{\lambda}{2N}
                    \left(
                        \alpha
                        - \frac{1}{\tau |\mathcal{\bar B}^+_q|}
                    \right)
                \cr
                &=
                    \frac{(1 + \lambda)}{2N}
                    \left(
                        \alpha
                        -
                        \frac{
                            \lambda
                        }{
                            \tau
                            (1 + \lambda)
                            |\mathcal{\bar B}^+_q|
                        }
                    \right)
                \cr
                &\leq
                    \frac{(1 + \lambda)}{2N}
                    \left(
                        \alpha
                        -
                        \frac{
                            \lambda
                        }{
                            \tau
                            (1 + \lambda)
                            (2N - 2)
                        }
                    \right)
            \end{split}
        \end{equation}
        
        \noindent
        
        When $\alpha < \lambda \cdot \left\{\tau(1+\lambda)(2N-2)\right\}^{-1}$, the overall gradient of MCL becomes negative and urges every data points to gather near the centroid of each cluster.
        So hyper-parameter $\alpha$ should be carefully selected to avoid the following condition:
        
        \begin{equation}
            \label{eq:contraction-condition}
            \alpha
            <
            \frac{
                \lambda
            }{
                \tau
                (1 + \lambda)
                (2N - 2)
            }
            .
        \end{equation}



    \subsection{MCL Hyper-parameter Search}
    
        In our experiment setting, contrastive learning related hyper-parameters follow SimCLR \cite{chen2020simple}, setting $\tau$ to 0.2 and $N$ to 1024, to keep our experiment tractable.
        We carefully selected the remaining MCL-related hyper-parameters, $\alpha$ and $\beta$, that meet 2 conditions from the previous subsection.
        To calculate exact value, we substitute the corresponding hyper-parameter value to Eq. \ref{eq:attraction-condition} and Eq. \ref{eq:contraction-condition}.
        We approximated $|\mathcal{\bar B}^+_q| \approx 2N/C = 51.2$, since the total number of classes $C$ in MCL with auxiliary task is $C = 10 \times 4 = 40$.
        Then, Eq. \ref{eq:attraction-condition} is substituted as follows:
        \begin{equation}
            \alpha < \frac{1}{\tau |\mathcal{\bar B}^+_q|} \approx 0.098
        \end{equation}
        The final value for $\alpha$ is set to 0.05 which meets both conditions in Eq. \ref{eq:attraction-condition} and Eq. \ref{eq:contraction-condition}.
        
        Remaining hyper-parameter $\beta$ is set to $ 1/(2\cdot\tau) = 2.5$. 
        By halving reprelling ratio, $\beta$ enables MCL to form hierarchical clusters by satisfying $0 < \alpha < \beta < 1/\tau$.
        As a side note, the $\beta$ value changes the formation of the cluster (hierarchical or not), however, does not affect a significant change in performance.

\section{Background: SupCLR}

        \cite{khosla2020supervised} proposed SupCLR, which is another \textit{task-specific} variant of SimCLR.
        Similar to MCL, SupCLR also leverages label information in the batch while training.
        Specifically, SupCLR additionally attracts all same label views from $\mathcal{\bar B}_{i}^{+}$ in the augmented batch $\mathcal{\bar B}$. 

        \begin{equation}
            p_\text{sup}(i, j) =
                \frac{
                    \text{exp} (
                        \text{sim}_{i, j}
                        /
                        \tau
                    )
                }{
                    \sum_{k = 1}^{2N}
                    \mathbbm{1}_{[k \neq i]}
                        \text{exp} (
                            \text{sim}_{i, k}
                            /
                            \tau
                        )
                }
            ,
        \end{equation}

        \begin{equation}
            \label{eq:SupCLR}
            \ell_\text{sup} (i) =
                \frac{1}{|\mathcal{\bar B}^{+}_{i}|}
                \sum_{(\bar{\bm{x}}_j, \bar y_j) \in \mathcal{\bar B}^{+}_{i}}
                    - \log p_\text{sup}(i, j)
        \end{equation}

        \noindent
        The final objective for SupCLR is to minimize Eq. \ref{eq:SupCLR} analogous to SimCLR:

        \begin{equation}
            \label{eq:loss-cl2}
            \mathcal{L}_\text{SupCLR} =
                \frac{1}{2N}
                \sum_{i=1}^{2N}
                    \ell_\text{sup} (i)
            .
        \end{equation}
        
        Since SupCLR forces all positive views to have a high similarity to the query view, the ability to distinguish data within the same class is diluted.
        In MCL, $\alpha$ in CCM suppresses positive samples having too high similarity with query view.
        Unlike SupCLR which does not repel positive views in a batch, MCL also repels positive views by a small ratio $\alpha$.
        By penalizing a small amount $\alpha$ to positive views, MCL can discern each data in the same label cluster.
        Such technique is analogous to the smoothing (LS) technique a widespread strategy to prevent over-confident prediction problems in general supervised classifiers, where $\alpha$ plays similar a role to smoothing-ratio in LS.
        As a side note, the test accuracy can be further improved by increasing the number of stochastic samples in the MCL or by increasing $\lambda$ in SPA.
        However, mentioned tactics decrease discrepancy between IND distribution and OOD distribution, undermining the AUROC score.
        Since test-accuracy is not the primary concern in anomaly detection, hyper-parameters and model components in MCL are set to the value that shows the best AUROC performance.
        

\begin{figure*}[ht!]
    \centering
    \includegraphics[width=1\textwidth]{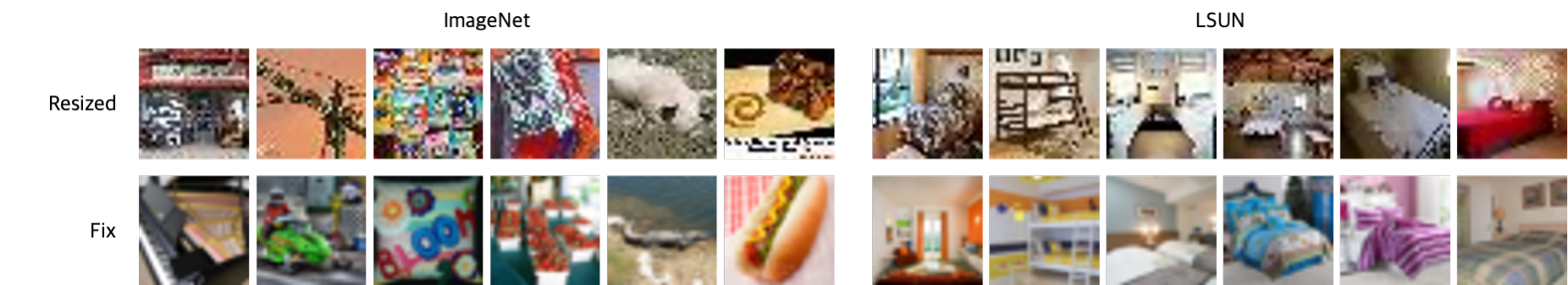}
    \caption{Samples from resized and fixed versions of ImageNet and LSUN for OOD detection. Because the resizing was done by subsampling, it caused an aliasing effect, resulting in unintended artifacts that made images almost indistinguishable even for humans. Fixed version used interpolation technique for the resizing.}
    \label{fig:appendix-dataset-sample}
\end{figure*}

\section{Expanded Experimental Results}
    \subsection{Dataset Description}
    
        We conduct experiments on various multi-class benchmark datasets: CIFAR-10, CIFAR-100 \cite{krizhevsky2009learning}, SVHN \cite{netzer2011reading}, Tiny-ImageNet \cite{deng2009imagenet}, and LSUN \cite{yu2015lsun}.
    
        CIFAR-10, CIFAR-100, Tiny-ImageNet dataset is a widely used multi-class  natural image (\textit{e.g.}, dog, tiger, plane) dataset with 10,100, and 200 classes label for each dataset.
        SVHN dataset stands for Street View House Numbers (SVHN), which consist of ten-digit classes.
        LSUN dataset contains millions of color images with 10 scene categories and 20 object categories.
        Since the size of each dataset differs, several data sets have to be resized to unify to fit the targeted size (32 by 32).
        \cite{liang2017enhancing} released resized version of Tiny-ImageNet and LSUN dataset, however, as can be seen in Figure \ref{fig:appendix-dataset-sample}, those version contains unintended artifacts produced during the resizing process.
        \cite{tack2020csi} handled the aforesaid problem and released a fixed version of those datasets dubbed Tiny-ImageNet(F) and LSUN(F).

    \subsection{Additional OOD Metrics}

        We provided an extended table for our experimental result with additional OOD performance metrics: FPR@95, AUPR

        \begin{itemize}
            \item \textbf{FPR@95:} The ROC curve is a graph where the x-axis and y-axis in a graph indicate FPR and TPR respectively. FPR@95 is a specific FPR point of the ROC curve when TPR is set to 95. Lower FPR indicates fewer false positives arises.
            \item \textbf{AUPR:} AUPR stands for \textit{area under the precision-recall curve} (PR-curve). In particular, two PR-curves are produced, from IND perspective PR-curve and OOD perspective PR-curve. Higher AUPR indicates high precision and recall value.
        \end{itemize}


    \subsection{Expanded Results}
    
        In this section, we provide experimental results with additional OOD metrics in Tab. \ref{tab:appendix-result-ablation-sei} and Tab. \ref{tab:appendix-result-ablation-overall}.
        From the perspective of AUROC, the performance improvement does not seem significant (difference around 1$\sim$2), however, from the perspective of FPR@95, about 20\% fewer false positives arises in MCL compared to the other existing methodologies.

    \subsection{Fine-tuning Setups}
        In the fine-tuning process, we adopt equivalent structure used in MCL with additional \textit{task-specific} layer on the top level.
        We trained 100 epochs using cosine annealing scheduler with  batch size 512 and learning rate 0.2.
        The objective for joint training \cite{winkens2020contrastive} is done by combining \textit{cross-entropy loss} and SimCLR loss (Eq. \ref{eq:loss-cl}) as follows:
        \begin{equation}
            \label{eq:loss-joint}
            \mathcal{L}_\text{joint} =
                \mathcal{L}_{\text{CE}}
                +
                \lambda \mathcal{L}_{\text{SimCLR}}
            ,
        \end{equation}
        where $\lambda$ denotes loss weight for \textit{cross-entropy loss}. We set $\lambda$ to 0.1, where we get the best performance from our $\lambda$ candidates. (0.01, 0.1, 1)
        We also fixed hyper-parameters related to contrastive learning following SimCLR, which include transformation $\mathcal{T}= \{ $color jittering, horizontal flip, grayscale, inception crop$\}$, the strength of color distortion to 0.5, in the case of joint training.

\section{Expanded Related Work}

    Anomaly detection, also termed OOD, novelty, or outlier detection, is a research area that aims to identify anomalies by distinguishing whether the test sample is drawn from in-distribution or not.
    Most of the recent methods in anomaly detection use deep-learning which can be branched into the following categories:

    \noindent
    \textit{Generative and hybrid models} use a generative model to measure the uncertainty of test data. The most common method is measuring reconstruction error, assuming that the model learns a proper mapping function that successfully reconstructs normal data samples with a very small reconstruction error \cite{oza2019c2ae,li2018anomaly,schlegl2017unsupervised}.
    Another methodology is to train the distribution of the normal data so that the model can assign a low likelihood to anomalous data \cite{zhang2020hybrid,gal2016dropout,malinin2018predictive,blundell2015weight}.

    \noindent
    \textit{Discriminative models }leverage information from trained classifiers such as \textit{maximum softmax probability} (MSP), or Mahalanobis distance of latent feature. \cite{hendrycks2016baseline} presented the most natural baseline for a discriminative model (classifier) which distinguishes anomalies by the predictive probability (MSP) of the classifier. \cite{lee2018simple,liang2017enhancing} proposed a post-processing method that adds a small adversarial like perturbation on the input.
    \cite{hendrycks2018deep} proposed a supervised OOD method by enforcing uniform distribution to anomalous data. More recently, Self-supervised Learning is spurring interest in anomaly detection, since access to OOD data in the most real-world scenario is quite unfeasible.
    \cite{golan2018deep}, one of the earlier works to identify the potential of SSL, proposed a simple but effective technique that aims to discriminate within in-distribution (IND) samples through 3 auxiliary tasks. (\textit{e.g.}, flip, rotation, and translation)
    Furthermore, \cite{hendrycks2019using} confirmed that using auxiliary tasks not only helps to determine OOD samples but also helps to defend against adversarial attacks. In more recent works, \cite{tack2020csi,winkens2020contrastive} used contrastive learning to get well-suited representation for anomaly detection.
    In \cite{tack2020csi}, particularly, the performance improvement was achieved by considering transforms known to be harmful as negative. (\textit{e.g.}, rotations, cutout)

\begin{table*}
    \centering
    \setlength{\tabcolsep}{9pt}
    \renewcommand{\arraystretch}{1.2}
    \begin{tabular}{c|c|l|c|cccc}

        \toprule

            \multirow{3}{*}{Methods} & \multirow{3}{*}{SEI} & \multirow{3}{*}{Model} & \multirow{3}{*}{Acc $\uparrow$} & \multicolumn{4}{c}{CIFAR-100} \\ \cline{5-8}
            & & & & \multirow{2}{*}{AUROC $\uparrow$} & \multirow{2}{*}{FPR@95 $\downarrow$} & AUPR $\uparrow$ & AUPR $\uparrow$ \\
            & & & & & & (IND) & (OOD) \\

        \midrule

            \multirow{4}{*}{MCL (w/o Aux)}

            & \multirow{2}{*}{-}
            & LF
            & 91.74 & 90.07 & 57.55 & 91.63 & 87.23 \\
            &
            & PH
            & 94.35 & 90.49 & 52.69 & 91.28 & 87.58 \\

            \cline{2-8}

            & \multirow{2}{*}{w-avg}
            & 8-way SEI + LF
            & 86.59&76.89 &70.77 &75.89 & 76.66\\
            &
            & 8-way SEI + PH
            & 83.96 & 71.17 &76.11 &69.64 & 71.59\\

        \midrule

            \multirow{4}{*}{MCL (w/o Aux) + DA}

            & \multirow{2}{*}{-}
            & LF
            & 91.78 & 90.00 & 57.74 & 91.55 & 87.17 \\
            &
            & PH
            & 92.55 & 90.07 & 54.93 & 91.17 & 87.89 \\

            \cline{2-8}

            & \multirow{2}{*}{w-avg}
            & 8-way SEI + LF
            & 93.49 & 92.65 & 50.46 & 93.00 & 89.36 \\
            &
            & 8-way SEI + PH
            & 94.70 & 92.08 & 44.94 & 92.92 & 90.55 \\

        \midrule

            \multirow{14}{*}{MCL}

            & \multirow{2}{*}{-}
            & LF
            & 90.04 & 91.00 & 51.56 & 92.30 & 88.62 \\
            &
            & PH
            & 94.03 & 91.12 & 48.16 & 91.80 & 89.24 \\

            \cline{2-8}

            & \multirow{4}{*}{avg}
            & 4-way SEI + LF
            & 91.12 & 91.98 & 49.27 & 93.30 & 89.46 \\
            &
            & 4-way SEI + PH
            & 94.68 & 93.20 & 38.24 & 93.83 & 91.73 \\
            &
            & 8-way SEI + LF
            & 91.61 & 92.14 & 47.37 & 93.41 & 89.67 \\
            &
            & 8-way SEI + PH
            & 94.74 & 93.37 & 36.23 & 93.89 & 92.04 \\

            \cline{2-8}

            & \multirow{4}{*}{max}
            & 4-way SEI + LF
            & 93.62 & 92.06 & 45.66 & 93.11 & 89.37 \\
            &
            & 4-way SEI + PH
            & 95.97 & 92.30 & 40.26 & 92.33 & 91.27 \\
            &
            & 8-way SEI + LF
            & 94.35 & 92.00 & 44.63 & 93.00 & 89.19 \\
            &
            & 8-way SEI + PH
            & 96.40 & 92.00 & 40.01 & 91.84 & 91.11 \\

            \cline{2-8}

            & \multirow{4}{*}{w-avg}
            & 4-way SEI + LF
            & 91.94 & 92.34 & 45.63 & 93.50 & 90.02 \\
            &
            & 4-way SEI + PH
            & 96.12 & 93.29 & 37.06 & 93.71 & 92.17 \\
            &
            & 8-way SEI + LF
            & 92.43 & 92.53 & 44.38 & 93.64 & 90.28 \\
            &
            & 8-way SEI + PH
            & \textbf{96.43} & \textbf{94.06} & \textbf{32.67} & \textbf{94.59} & \textbf{93.21} \\

        \bottomrule

    \end{tabular}
    \caption{Ablation studies for SEI. Backbone architecture is fixed as ResNet-34. Latent feature, projection head, auxiliary task, and data augmentation is abbreviated to LF, PH, Aux, and DA respectively. Symbol - indicates a model without SEI.}
    \label{tab:appendix-result-ablation-sei2}
\end{table*}

\begin{table*}[p]
    \centering
    \setlength{\tabcolsep}{8pt}
    \renewcommand{\arraystretch}{1.2}
    \begin{tabular}{c|l|c|ccccc}

        \toprule

            \multirow{2}{*}{Backbone} & \multirow{2}{*}{Model} & \multirow{2}{*}{Acc $\uparrow$} & \multicolumn{4}{c}{CIFAR-100} \\ \cline{4-7}
            & & & AUROC $\uparrow$ & FPR@95 $\downarrow$ & AUPR (IND) $\uparrow$ & AUPR (OOD) $\uparrow$ \\

        \midrule

            \multirow{7}{*}{ResNet-34}

            & MCL (w/o SPA, w/o aux)
            & 91.41 & 85.03 & 60.70 & 84.26 & 83.65 \\

            & MCL (w/o aux)
            & 94.35 & 90.49 & 52.69 & 91.28 & 87.58 \\

            & MCL (w/o aux) + DA
            & 92.55 & 90.07 & 54.93 & 91.17 & 87.98 \\

            & MCL (w/o aux) + DA + SEI
            & 94.70 & 92.08 & 44.94 & 92.92 & 90.55 \\

            & MCL
            & 94.03 & 91.12 & 48.16 & 91.80 & 89.24 \\

            & MCL + fine-tuning
            & 94.22 & 88.89 & 62.84 & 90.20 & 84.70 \\

            & MCL + SEI
            & \textbf{96.43} & \textbf{94.06} & \textbf{32.67} & \textbf{94.59} & \textbf{93.21} \\

        \bottomrule

    \end{tabular}

    \caption{Ablation study for MCL with additional OOD metrics.}
    \label{tab:appendix-result-ablation-overall}
\end{table*}

\begin{figure*}[p]
    \begin{subfigure}{.49\linewidth}
        \centering
        \includegraphics[width=.95\linewidth]{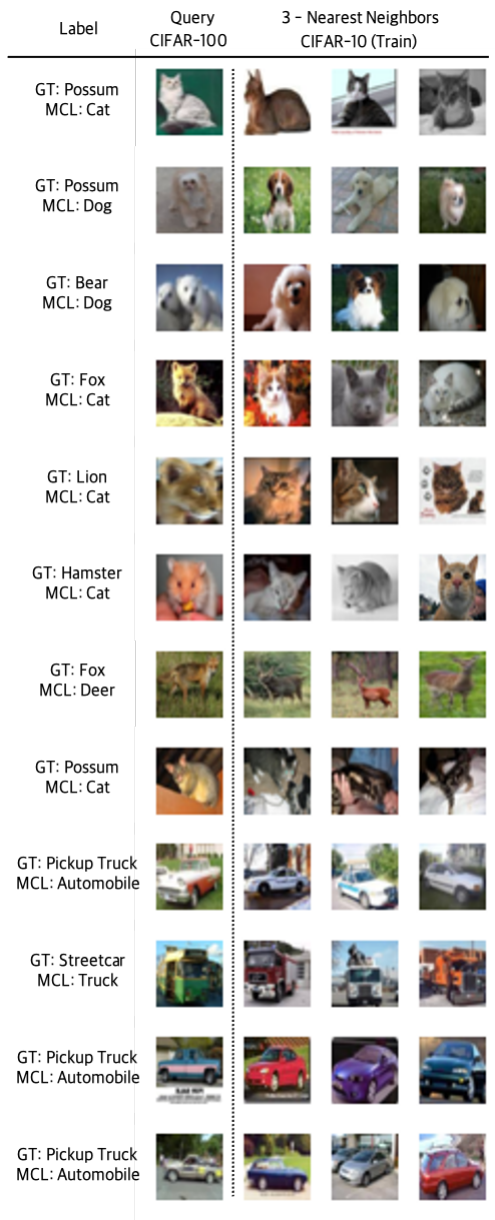}
        \caption{More OOD samples with high IND scores.}
        \label{fig:appendix-case-ood}
    \end{subfigure}
    \begin{subfigure}{.49\linewidth}
        \centering
          \includegraphics[width=.95\linewidth]{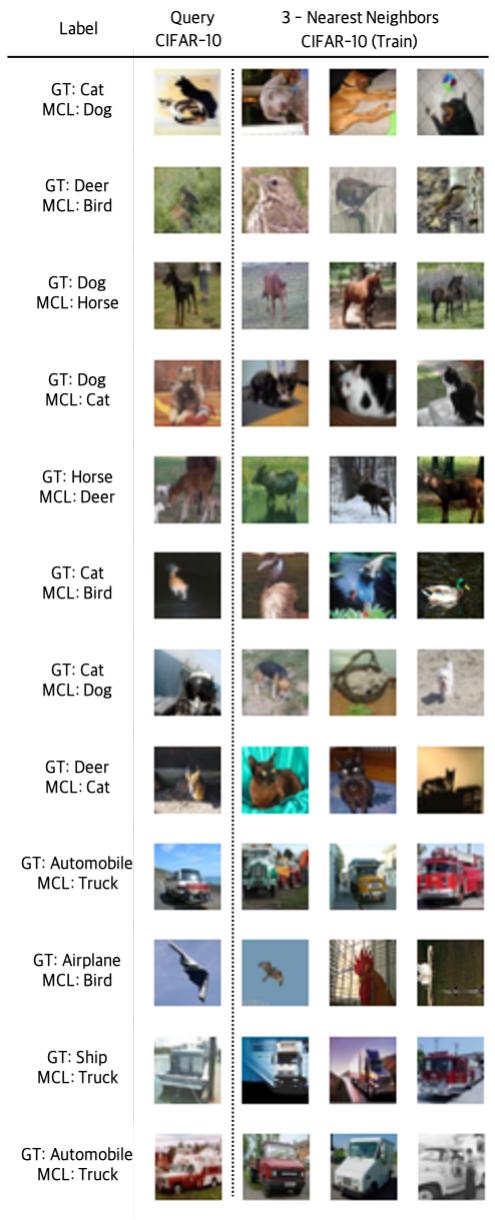}
        \caption{More wrongly classified samples from IND.}
        \label{fig:appendix-case-ind}
    \end{subfigure}
    \caption{More results on CIFAR-10 with CIFAR-100 as OOD. GT stands for the ground truth label. Each row shows a query image on the second column, followed by the top 3 most similar images of train data. Quite an amount of samples are either wrong, ambiguous, or vague. (a) Query images are from OOD, but the model took as an IND sample and classified with high confidence. (b) Wrongly classified samples from in-domain classification.}
    \label{fig:appendix-case}
\end{figure*}

\end{document}